\documentclass[lettersize,journal]{IEEEtran}
\usepackage{amsmath,amsfonts}
\usepackage{algorithmic}
\usepackage{algorithm}
\usepackage{array}
\usepackage[caption=false,font=normalsize,labelfont=sf,textfont=sf]{subfig}
\usepackage{textcomp}
\usepackage{stfloats}
\usepackage{url}
\usepackage{verbatim}
\usepackage{graphicx}
\usepackage{cite}
\usepackage{amssymb}
\usepackage{booktabs}
\usepackage{multirow}
\usepackage{ulem}
\usepackage{color}
\usepackage{pifont}
\usepackage{balance}
\usepackage{colortbl}
\usepackage{xcolor}
\usepackage{makecell}
\usepackage{amsmath,amssymb,mathrsfs}
\usepackage{bm}
\newcommand{\cmark}{\ding{51}}

\usepackage{balance}
\usepackage{bbding} 
\usepackage{multirow}
\definecolor{newcolor}{rgb}{.8,.349,.1}

\begin{document}

\title{MonoDiffusion: Self-Supervised Monocular Depth Estimation Using Diffusion Model}

\author{Shuwei Shao, Zhongcai Pei, Weihai Chen$^{*}$, Dingchi Sun, Peter C.Y.Chen and Zhengguo Li, \textit{Fellow, IEEE}
\thanks{This work was supported by the National Natural Science Foundation of China under grant 61620106012. }
\thanks{Shuwei Shao, Zhongcai Pei, Weihai Chen and Dingchi Sun are with the School of Automation Science and Electrical Engineering, Beihang University, Beijing, China. (email: swshao@buaa.edu.cn, peizc@buaa.edu.cn, whchen@buaa.edu.cn, sdc@nuaa.edu.cn)}
\thanks{Peter C.Y.Chen is with the Department of Mechanical Engineering, National University of Singapore, Singapore. (e-mail: mpechenp@nus.edu.sg)}
\thanks{Zhengguo Li is with the SRO department, Institute for Infocomm Research, 1 Fusionopolis Way, Singapore. (e-mail: ezgli@i2r.a-star.edu.sg)}
\thanks{*(corresponding author: Weihai Chen.)}
}

\markboth{xxx, VOL. XX, NO. XX, XXXX 2023}%
{Shao \MakeLowercase{\textit{et al.}}: MonoDiffusion: Self-Supervised Monocular Depth Estimation \\ Using Diffusion Model}


\maketitle

\begin{abstract}
Over the past few years, self-supervised monocular depth estimation that does not depend on ground-truth during the training phase has received widespread attention. Most efforts focus on designing different types of network architectures and loss functions or handling edge cases,~\textit{e.g.}, occlusion and dynamic objects. In this work, we introduce a novel self-supervised depth estimation framework, dubbed MonoDiffusion, by formulating it as an iterative denoising process. Because the depth ground-truth is unavailable in the training phase, we develop a pseudo ground-truth diffusion process to assist the diffusion in MonoDiffusion. The pseudo ground-truth diffusion gradually adds noise to the depth map generated by a pre-trained teacher model. Moreover, the teacher model allows applying a distillation loss to guide the denoised depth. Further, we develop a masked visual condition mechanism to enhance the denoising ability of model. Extensive experiments are conducted on the KITTI and Make3D datasets and the proposed MonoDiffusion outperforms prior state-of-the-art competitors. The source code will be available at \url{https://github.com/ShuweiShao/MonoDiffusion}. 
\end{abstract}


\section{Introduction}

\IEEEPARstart{M}onocular depth estimation (MDE) is one of the fundamental tasks in the computer vision community, with many applications,~\textit{e.g.}, 3D reconstruction, scene understanding and autonomous driving~\cite{izadi2011kinectfusion,chen2019towards,natan2022end}. Recently, learning-based approaches~\cite{eigen2014depth,zhou2017unsupervised,yin2019enforcing,wang2023planedepth} have achieved remarkable advances, where the full supervised MDE~\cite{lee2019big,bhat2021adabins,kim2018deep,shao2023urcdc} shows higher accuracy due to the available depth ground-truth. Nonetheless, the ground-truth is hard to acquire because of expensive hardware sensors, sensor noise, limited operating capabilities, etc.

Self-supervised MDE~\cite{zhou2017unsupervised,ye2021unsupervised,peng2021excavating,karpov2022exploring} has been proposed as a promising alternative by formulating the MDE as a task of novel view synthesis. When leveraging stereo image pairs, the motion of the camera is available, allowing for the use of a separate depth estimation network in the training phase. When training on monocular videos, an additional pose network is necessary to estimate the camera motion. Nevertheless, self-supervised MDE that relies solely on monocular videos are preferred because of the difficulty in collecting stereo data, for example, intricate configurations and data processing. To boost the performance, following methods developed improved loss functions~\cite{godard2019digging} or leveraged semantic information~\cite{jung2021fine,klingner2020self} to solve the dynamic objects and occlusion. A recent work Lite-Mono~\cite{zhang2023lite} integrated convolutional neural network (CNN) and Transformer to design more powerful network architecture.

\begin{figure}[!]
	\centering
	\includegraphics[width=0.99\linewidth]{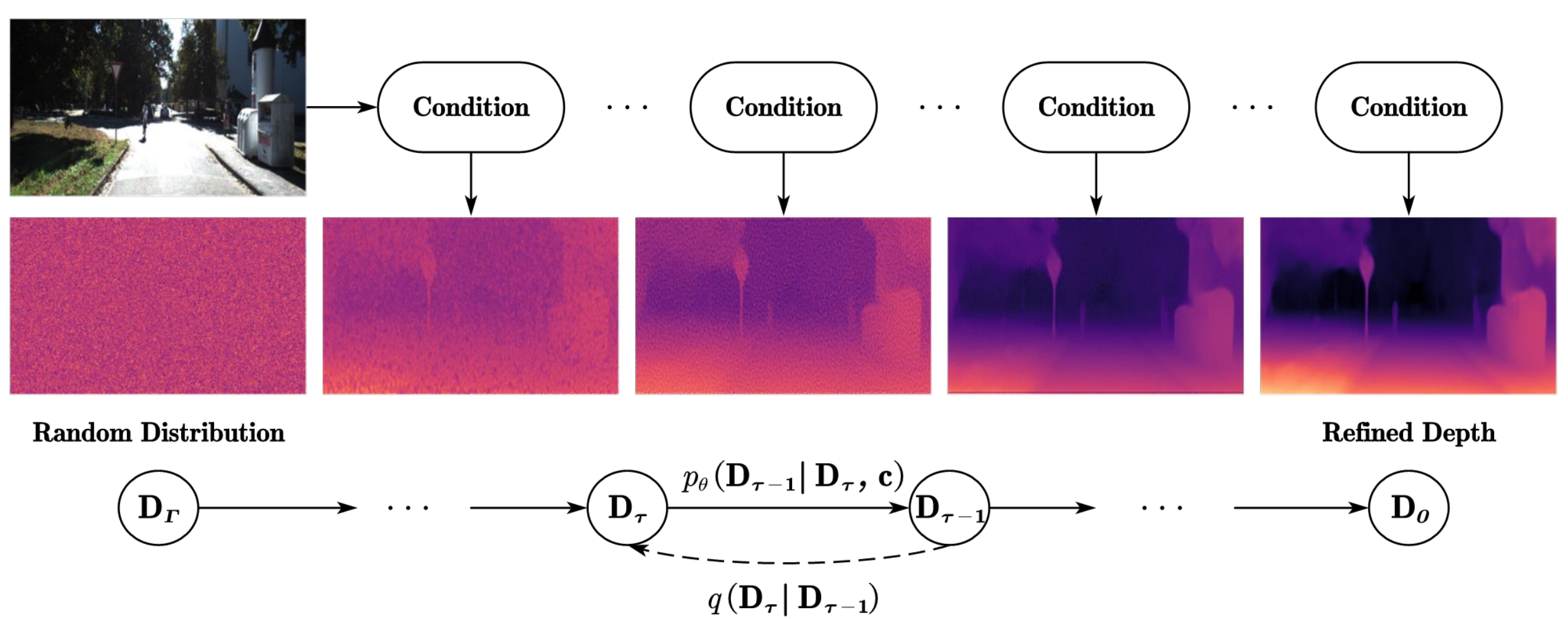}
	\caption{\textbf{Illustration of the denoising process guided by visual conditions.}}
	\label{Fig1}
\end{figure}
In this paper, we reformulate self-supervised MDE from the perspective of iterative denoising starting with a random depth distribution, as presented in Fig.~\ref{Fig1}. Diffusion models have attracted a widespread attention due to its efficacy in generative tasks~\cite{hoogeboom2022equivariant,trippe2022diffusion}, detection~\cite{chen2022diffusiondet} and segmentation~\cite{chen2022generalist}. More recently, Saxena~\textit{et al.}~\cite{saxena2023monocular} and Duan~\textit{et al.}~\cite{duan2023diffusiondepth} applied the diffusion model to MDE in a full supervised setting. However, the lack of depth ground-truth in self-supervised MDE poses a severe challenge to the diffusion process that requires ground-truth in diffusion model. 

We propose MonoDiffusion, which takes as input a random depth distribution and progressively refines it through multiple denoising steps under the guidance of visual conditions. Since the depth ground-truth is not available in the training phase, we introduce a pseudo ground-truth diffusion process to assist the diffusion in MonoDiffusion. Specifically, the pseudo ground-truth diffusion gradually appends noise to the depth generated by a pre-trained teacher model, which we refer to as the pseudo ground-truth. As a by-product, the teacher model allows us to impose a knowledge distillation loss on the denoised depth to improve the results. In order to alleviate the negative impact of depth error in pseudo ground-truth, we apply a multi-view check filter~\cite{liu2023self} to filter out erroneous depth. Furthermore, we develop a masked visual condition mechanism to enhance the denoising ability of MonoDiffusion, inspired by the success of combining diffusion model with masked image modeling~\cite{gao2023masked,wei2023diffusion}. The difference is that our target to reconstruction is the denoised depth rather than the RGB image. 

To summarize, our contributions are listed as follows:

\begin{itemize}
	\item We propose a novel framework, dubbed MonoDiffusion, for self-supervised monocular depth estimation by regarding it as an iterative denoising process.
	
	\item We introduce a pseudo ground-truth diffusion process to assist the diffusion in MonoDiffusion and a masked visual condition mechanism to enhance its denoising ability.
	
	\item Extensive experiments show that the proposed MonoDiffusion surpasses previous state-of-the-art competitors on the KITTI~\cite{geiger2012we} and Make3D~\cite{saxena2008make3d} datasets.
\end{itemize}
\begin{figure*}[htb!]
	\centering
	\includegraphics[width=0.93\linewidth]{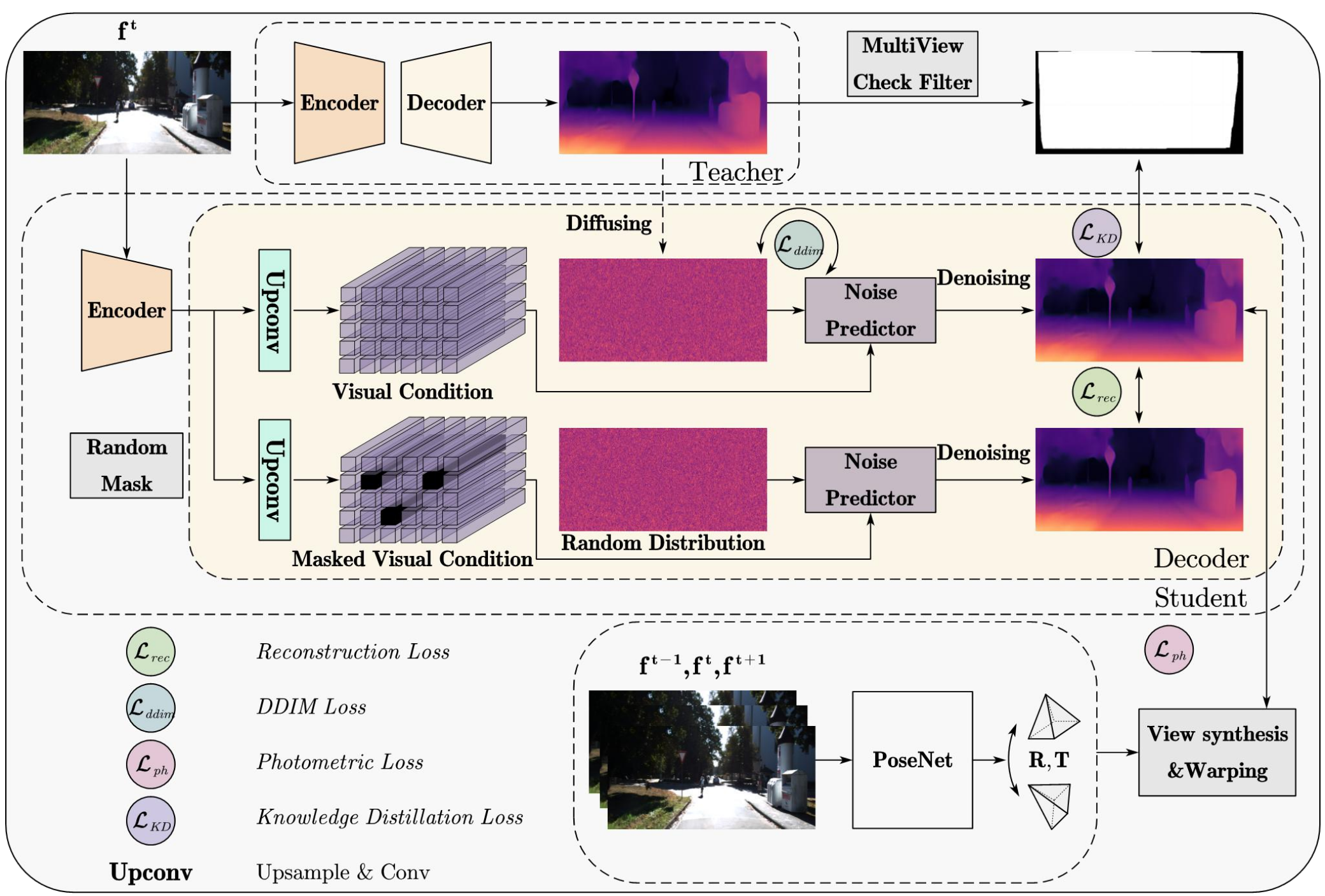}
	\caption{\textbf{An overview of the proposed MonoDiffusion}. During the training phase, MonoDiffusion involves an additional teacher model to assist the diffusion. The teacher model is self-supervised pre-trained based on Lite-Mono~\cite{zhang2023lite} and will be discarded once the training is completed.}
	\label{Fig2}
\end{figure*}


\section{Related Work}
\subsection{Monocular Depth Estimation}
MDE aims to predict depth from a single image, which is an ill-posed problem because there are infinitely many 3D scenes that can be projected onto the same 2D image. The relevant methods are broadly categorized into two groups.

\textbf{Supervised depth estimation} uses depth ground-truth to establish supervision and has achieved outstanding performance. Eigen~\textit{et al.}~\cite{eigen2014depth} introduced the first attempt of using CNN to perform multi-scale depth estimation. Then, Laina~\textit{et al.}~\cite{laina2016deeper} utilized the residual CNN~\cite{he2016deep} to enable network optimization to be easier. Cao~\textit{et al.}~\cite{cao2017estimating} and Fu~\textit{et al.}~\cite{fu2018deep} discretized the full depth range into multiple intervals and chose the optimal interval as depth estimate. Yuan~\textit{et al.}~\cite{Yuan_2022_CVPR} developed neural window fully-connected conditional random fields (CRFs) to reduce the computation in conventional CRFs. Shao~\textit{et al.}~\cite{shao2023urcdc} introduced the cross-distillation paradigm to integrate strengths from CNN and Transformer. Liu~\textit{et al.}~\cite{liu2023va} enforced the first-order variational constraint to regularize depth map. However, these methods rely heavily on the quality and quantity of depth ground-truth while the proposed MonoDiffusion does not have such a requirement.


\textbf{Self-supervised depth estimation} does not require costly depth ground-truth and employ stereo image pairs or adjacent frames in a video to generate the supervisory signal. As one of the pioneering works, Zhou~\textit{et al.}~\cite{zhou2017unsupervised} trained a depth network and a pose network to predict depth and 6-degree of freedom (DoF) pose, which are then utilized to perform view synthesis during training. To handle edge cases such as dynamic objects and occlusion, they proposed an explainability mask to remove these regions. Godard~\textit{et al.}~\cite{godard2019digging} further introduced an auto-masking technique and a per-pixel minimum re-projection loss to better handle dynamic objects and occlusion. Bian~\textit{et al.}~\cite{bian2019depth} devised a geometry consistency loss for scale-consistent depth and pose predictions. Johnston~\textit{et al.}~\cite{johnston2020self} and Liu~\textit{et al.}~\cite{liu2023self} utilized the discrete disparity prediction and self-reference distillation to boost performance, respectively. Zhang~\textit{et al.}~\cite{zhang2023lite} developed an efficient hybrid architecture by combining CNN and Transformer. In contrast, we introduce the diffusion model and draw on its strong generative capability to produce high-quality depth predictions.

   
\subsection{Diffusion Model}
Diffusion models have received widespread attention due to its effectiveness in image generation~\cite{ho2020denoising,song2020score,dhariwal2021diffusion}. Nevertheless, its potential in downstream tasks has largely been unexplored. Fortunately, Song~\textit{et al.}~\cite{song2020denoising} improved the denoising process to make inference steps more affordable for these tasks.~\cite{wolleb2022diffusion,brempong2022denoising,baranchuk2021label} extended diffusion model in image segmentation and Chen~\textit{et al.}~\cite{chen2022diffusiondet} leveraged diffusion model to generate detection box proposals. There are also attempts at applying diffusion models to MDE~\cite{saxena2023monocular,duan2023diffusiondepth}, but they focus on a fully supervised setting. To alleviate the performance degradation induced by diffusing on sparse ground-truth, Saxena~\textit{et al.}~\cite{saxena2023monocular} and Duan~\textit{et al.}~\cite{duan2023diffusiondepth} both proposed to diffuse on the denoised output of network to assist training. Self-supervised MDE with the diffusion model is more challenging due to the lack of depth ground-truth. In addition, we find that diffusing on the denoised output as done by these two methods does not work well for self-supervised MDE.

\section{Methodology}
In this section, we first introduce the preliminary knowledge of diffusion model and self-supervised MDE. Then, we elaborate on the proposed MonoDiffusion, namely, depth estimation as denoising, pseudo ground-truth diffusion and masked visual condition. Finally, we demonstrate the overall architecture and loss. An overview of the whole framework is shown in Fig.~\ref{Fig2}.

\subsection{Preliminaries}
\textbf{Diffusion models}, for example,~\cite{song2020denoising,ho2020denoising,sohl2015deep}, belong to the category of latent variable models and widely employed in generative tasks. In practice, they are trained to denoise images blurred with Gaussian noise and reverse the diffusion process $q\left( {{\textbf{x}_\tau}\left| {{\textbf{x}_0}} \right.} \right)$, which involves the iterative addition of noise to a desired image distribution $\textbf{x}_0$ and allows acquiring the latent noisy sample $\textbf{x}_\tau$. Mathematically,
\begin{equation}
	q\left( {{\textbf{x}_\tau}\left| {{\textbf{x}_0}} \right.} \right): = \mathcal{N}\left( {{\textbf{x}_\tau}\left| {\sqrt {{{\overline \alpha  }_\tau}} {\textbf{x}_0},(1 - {{\overline \alpha  }_\tau})\textbf{I}} \right.} \right),
\end{equation}
with
\begin{equation}
	{\overline \alpha  _\tau }: = \prod\nolimits_{n = 0}^\tau  {{\alpha _\tau }}  = \prod\nolimits_{n = 0}^\tau  {\left( {1 - {\beta _n}} \right)},
\end{equation}
and
\begin{equation}
{\textbf{x}_\tau } = \sqrt {{{\overline \alpha  }_\tau }} {\textbf{x}_0} + \sqrt {1 - {{\overline \alpha  }_\tau }} \epsilon , \quad \epsilon  \sim \mathcal{N}\left( {0,1} \right), 
\end{equation}
where $\tau$ consists of \textbf{T} steps, ${\beta _n}$ stands for the noise variance schedule as DDPM~\cite{ho2020denoising} and $\mathcal{N}$ stands for the Gaussian noise. 

For the denoising process, a noise predictor ${\epsilon_\theta }\left( {{\textbf{x}_\tau},\tau } \right)$ learns to reverse the diffusion process and recover $\textbf{x}_0$. Each denoising step is approximated by a Gaussian distribution, 
\begin{equation}
	{p_\theta }\left( {{\textbf{x}_{\tau  - 1}}\left| {{\textbf{x}_\tau }} \right.} \right) : = \mathcal{N}\left( {{\textbf{x}_{\tau  - 1}};{\mu _\theta }\left( {{\textbf{x}_\tau },\tau } \right), \sigma _\tau ^2\textbf{I}}  \right),
\end{equation}
where ${\mu _\theta }\left( {{\textbf{x}_\tau },\tau } \right)$ is acquired by the linear combination of ${\textbf{x}_\tau }$ and predicted noise of ${\epsilon_\theta }\left( {{\textbf{x}_\tau},\tau } \right)$ and $\sigma _\tau ^2$ denotes the transition variance.~\\ 

\textbf{Self-supervised MDE} frames the depth estimation as a task of novel view synthesis and typically requires two networks, a depth network and an additional pose network. To achieve supervision, the predicted depth map is utilized to back-project each pixel into 3D camera space with the camera intrinsic. The acquired 3D point cloud is then transformed to another view utilizing the predicted relative pose. Suppose a target frame ${\textbf{f}^t}$ and a source frame ${\textbf{f}^s}$, the procedure is defined as 
\begin{equation}
	{\textbf{p}^{t \to s}} = \textbf{K}{\textbf{M}^{t \to s}}{\textbf{D}^t}\left( \textbf{p}  \right){\textbf{K}^{ - 1}}{\textbf{p}^t}, \label{eq4}
\end{equation}
where $\textbf{p}^{t \to s}$ denotes the mapped pixel coordinate from target view $t$ to source view $s$, $\textbf{p}^{t}$ denotes the pixel coordinate in the target view, $\textbf{K}$ denotes the camera intrinsic, $\textbf{M}^{t \to s}$ denotes the relative pose from target view to source view, and ${\textbf{D}^t}$ denotes the depth map of target frame. Next, we can obtain a warped frame ${\textbf{f}^{s \to t}}$ via
\begin{equation}
	{\textbf{f}^{s \to t}}(\textbf{p}) = {\textbf{f}^s}\left\langle {{\textbf{p}^{t \to s}}} \right\rangle,
\end{equation}
where $\left\langle  \cdot  \right\rangle$ denotes the warping operation~\cite{jaderberg2015spatial}. The appearance difference between ${\textbf{f}^{s \to t}}$ and ${\textbf{f}^t}$ is used to supervise the whole framework. As in most works,~\textit{e.g.}, \cite{godard2019digging, yang2020d3vo},  a combination of L1 loss and structural similarity (SSIM) term~\cite{wang2004image} measures the dissimilarity in appearance, written as
\begin{equation}\begin{split} {{\mathcal{L}}_{ph}}=\sum_{\textbf{p}}\kappa \displaystyle{\frac{1 - {\rm{SSIM}}\left( {{{\textbf{f}^{t}}\left( {\textbf{p}} \right)},{{\textbf{f}^{s \to t}}\left( \textbf{p} \right)}} \right)}{2}}\\ +\sum_{\textbf{p}}\left( 1-\kappa  \right){{\left\| {{{\textbf{f}^{t}}\left( {\textbf{p}} \right)} - {{\textbf{f}^{s \to t}}\left( \textbf{p} \right)}} \right\|}_{1}}, \end{split} \label{eq6} \end{equation} 
where ${\mathcal{L}}_{ph}$ is referred to as the photometric loss and $\kappa$ is set to 0.85.

\subsection{MonoDiffusion}

\textbf{Depth estimation as denoising.} Given the target frame ${\textbf{f}^t}$, a standard formulation for MDE is $p_\theta\left( {\textbf{D}^t \left| {\textbf{f}^t} \right.} \right)$. By contrast, we reframe the MDE as an iterative denoising process where visual conditions guide the refinement of random depth distribution $\textbf{D}_\tau$ into a depth estimate,
\begin{equation}
	{p_\theta }\left( {{\textbf{D}_{\tau  - 1}}\left| {{\textbf{D}_\tau },\textbf{{c}}} \right.} \right): = \mathcal{N}\left( {{\textbf{D}_{\tau  - 1}};{{\mu _\theta }}\left( {{\textbf{D}_\tau },\tau ,\textbf{c}} \right),\sigma _\tau ^2\textbf{I}} \right),
\end{equation}
where $\textbf{c}$ stands for the conditions.
To speed up the inference process, we employ an improved inference process from~\cite{song2020denoising}, where $\sigma _\tau ^2\textbf{I}$ is set to 0 to allow the output to be deterministic.~\\

\textbf{Pseudo ground-truth diffusion.} As described above, the diffusion process in diffusion model iteratively appends noise to a desired distribution, i.e., ground-truth. Unfortunately, the lack of depth ground-truth for self-supervised MDE makes it hard to perform. To address the issue, we introduce a pseudo ground-truth diffusion process. More specifically, we pre-train a teacher model based on Lite-Mono~\cite{zhang2023lite}, a well-behaved self-supervised monocular depth estimator. Then, the diffusion is performed on the depth generated by our teacher model, which we referred to as the pseudo ground-truth ${\textbf{D}_{pseudo}}\left( \textbf{p} \right)$, 
\begin{equation}
	q\left( {{\textbf{D}_\tau}\left| {{\textbf{D}_{pseudo}}} \right.} \right): = \mathcal{N}\left( {{\textbf{D}_\tau}\left| {\sqrt {{{\overline \alpha  }_\tau}} {\textbf{D}_{pseudo}},(1 - {{\overline \alpha  }_\tau})\textbf{I}} \right.} \right).
\end{equation}
The teacher model will be discarded once the training is completed. Compared with diffusion on the denoised output~\cite{saxena2023monocular,duan2023diffusiondepth}, the pseudo ground-truth process is able to avoid the noisy denoised output at the early training stage from deteriorating the entire training process. As a by-product, the teacher model allows imposing a knowledge distillation loss on the denoised depth. The knowledge distillation paradigm distills knowledge from the teacher model to a student model, which is beneficial to improve the results~\cite{shao2023urcdc}. To mitigate the adverse impact of depth error in the pseudo ground-truth, we apply a multi-view check  filter~\cite{liu2023self} to filter out erroneous depth. The knowledge distillation loss is thus defined as
\begin{equation} 
	{\mathcal{L}_{KD}} = \sum\limits_\textbf{p} {\Phi (\textbf{p}) \odot \left( {{\textbf{D}^t}\left( \textbf{p} \right) - {\textbf{D}_{pseudo}}\left( \textbf{p} \right)} \right)},
\end{equation}
where $\Phi (\textbf{p})$ stands for the multi-view check filter, $\odot$ denotes the element-wise multiplication. The core principle of multi-view check is similar to that of the novel view synthesis. We draw on Eq.~\ref{eq4} to get each projected point ${\textbf{p} ^{t \to s}}$ in the source view and  acquire a warped depth map,
\begin{equation}
	{\textbf{D}^{s \to t}}(\textbf{p}) = {\textbf{D}^s}\left\langle {{\textbf{p}^{t \to s}}} \right\rangle,
\end{equation}
where $\textbf{D}^s$ stands for the depth map of source frame. Similar to the projection procedure above, we leverage $\textbf{D}^s$ to perform a reprojection procedure to acquire each reprojected 2D point ${\widetilde{\textbf{p}}^{s \to t}}$ and depth map ${\widetilde{\textbf{D}}^{s \to t}}(\textbf{p})$ in the target view. Thereafter, a reprojection error $e_{repro}$ and a geometry error $e_{geo}$ are defined as
\begin{equation}
	{e_{repro}} = {\left\| {{\widetilde{\textbf{p}}^{s \to t}} - {\textbf{p}^t}} \right\|_2},
\end{equation}
\begin{equation}
	{e_{geo}} = \frac{{\left| {{{\widetilde {\textbf{D}}}^{s \to t}}\left( \textbf{p} \right) - {\textbf{D}^t}\left( \textbf{p} \right)} \right|}}{{{\textbf{D}^t}\left( \textbf{p} \right)}}, 
\end{equation}
The determination of valid pixels for multi-view check filter is achieved by
\begin{equation}
	\left\{ \textbf{p} \right\} = \left\{ {\textbf{p}\left| {{e_{repro}} < a{{\overline e }_{repro}},{e_{geo}} < b{{\overline e }_{geo}}} \right.} \right\}, \label{eq12}
\end{equation}
where ${{\overline e }_{repro}}$ and ${{\overline e }_{geo}}$ are the average over all pixels and $a$ and $b$ are set to 4 based on~\cite{liu2023self}.~\\

\textbf{Masked visual condition.} Inspired by the success of coupling diffusion model with masked image modeling~\cite{gao2023masked,wei2023diffusion}, we develop a masked visual condition mechanism to further enhance the denoising ability of MonoDiffusion. For multi-scale feature tokens propagated by the encoder, we generate random masks to remove part of the tokens from these feature tokens. To prevent the information leakage, these masks are shared otherwise masked information may be easily borrowed from feature tokens of different resolutions. More specifically, we generate a mask with the highest resolution, and leverage nearest neighbor interpolation to acquire a pyramid of masks. The masked feature tokens are aggregated into masked visual conditions through learnable layers composed of $3 \times 3$ convolutions and upsampling layers. Different from~\cite{gao2023masked,wei2023diffusion}, we make use of the masked visual conditions to reconstruct the denoised depth map guided by the complete visual conditions, instead of the RGB image. The reconstruction loss is defined as
\begin{equation}
	{\mathcal{L}_{rec}} = \sum\limits_\textbf{p} {\left| {{{\widehat {\textbf{D}}}^t}\left(\textbf{p} \right) - {\textbf{D}^t}\left( \textbf{p} \right)} \right|},
\end{equation}
where ${{\widehat {\textbf{D}}}^t}$ denotes the denoised depth map using masked visual conditions.

\begin{table}[htb]
	\begin{center}
		\renewcommand{\arraystretch}{1.2}
		\resizebox{1.0\columnwidth}{!}{\begin{tabular}{c|c|c|c}
				\hline
				\textbf{Layers} & \textbf{Channels} & \textbf{Input} & \textbf{Activation} \\
				\hline
				\hline
				mask 2     & 128 & encoder 2             & None \\
				upconv 2 & 128 & mask 2                   & ELU  \\
				upconv ${2_{skip}}$ & 144 & ↑upconv 2, encoder 1 & ELU  \\
				mask 1     & 64  & encoder 1            & None \\
				upconv 1 & 64  & mask 1                   & ELU \\
				upconv ${1_{skip}}$ & 88  & ↑upconv, encoder 0 & ELU \\
				mask 0     & 40  & encoder 0              & None \\
				upconv 0 & 40  & mask 0                   & ELU \\
				upconv ${0_{skip}}$ & 24  & ↑upconv0              & ELU \\
				NE  & 1   & random noise    & None \\
				TE   & 1   & scheduler timesteps        & None \\
				NP & 16  & upconv ${0_{skip}}$, NE, TE            & None \\
				\hline
			\end{tabular}
		}
	\end{center}
	\caption{\textbf{The decoder architecture of student depth network}. The kernel size and stride of convolution are $3 \times 3$ and 1. ``$\uparrow$'' denotes the $2 \times 2$ bi-linear upsampling. The subscript ``skip" denotes the skip connection. The NE, TE and NP denote noise embedding, time embedding and noise predictor, respectively. } 
	\label{table1}
\end{table}

\subsection{Network Architecture and Overall Loss}
Both \textbf{depth networks} adopt the prevalent encoder-decoder architecture. The teacher model is from~\cite{zhang2023lite}, which leverages Lite-Mono-8M as the encoder. On the other hand, the student model uses two encoders from Lite-Mono family in different settings, Lite-Mono and Lite-Mono-8M. Its decoder is adapted from~\cite{zhang2023lite} and multi-scale feature tokens from the encoder are aggregated into visual conditions through $3 \times 3$ convolutional layers and upsampling layers. The diffusion-denoising process is performed at the full resolution. The deployed noise predictor is lightweight, only involving an embedding layer and three $3 \times 3$ convolutional layers. The detailed decoder architecture of student depth network is presented in Table~\ref{table1}.

The \textbf{pose network } design is same to prior works~\cite{godard2019digging,zhang2023lite}, where a pre-trained ResNet18~\cite{he2016deep} is used as the encoder and four convolutional layers are used as the decoder. It receives a stacked pair of images and estimate a 6-DOF relative pose between them.

\begin{table*}[ht!]
	\begin{center}
		\scalebox{1.0}{
			\begin{tabular}{ccc|cccc|ccc|c}
				\hline
				\multirow{2}*{Method}& \multirow{2}*{Year}&
				\multirow{2}*{Data}&
				\multicolumn{4}{c|}{Depth Error ($\downarrow$)}&
				\multicolumn{3}{c|}{Depth Accuracy ($\uparrow$)}&
				\multicolumn{1}{c}{Model Size ($\downarrow$)}\\				
				\cline{4-11}
				&&& Abs Rel & Sq Rel & RMSE & RMSE log &$\delta <1.25$ &$\delta <1.25^2$&$\delta <1.25^3$&Params. \\
				\cline{1-11}
				GeoNet~\cite{yin2018geonet}&2018&M&0.149&1.060&5.567&0.226&0.796&0.935&0.975&31.6M\\
				DDVO~\cite{wang2018learning}&2018&M&0.151&1.257&5.583&0.228&0.810&0.936&0.974&28.1M\\
				Monodepth2-Res18~\cite{godard2019digging}&2019&M&0.115&0.903&4.863&0.193&0.877&0.959&0.981&14.3M\\
				Monodepth2-Res50~\cite{godard2019digging}&2019&M&0.110&0.831&4.642&0.187&0.883&0.962&0.982&32.5M\\
				SGDepth~\cite{klingner2020self}&2020&M+Se&0.113&0.835&4.693&0.191&0.879&0.961&0.981&16.3M\\
				Johnston~\textit{et al.}~\cite{johnston2020self}&2020&M&0.111&0.941&4.817&0.189&0.885&0.961&0.981&14.3M+\\
				CADepth-Res18~\cite{yan2021channel}&2021&M&0.110&0.812&4.686&0.187&0.882&0.962&\underline{0.983}&18.8M\\
				HR-Depth~\cite{lyu2021hr}&2021&M&0.109&0.792&4.632&0.185&0.884&0.962&\underline{0.983}&14.7M\\
				Lite-HR-Depth~\cite{lyu2021hr}&2021&M&0.116&0.845&4.841&0.190&0.866&0.957&0.982&3.1M\\
				R-MSFM3~\cite{zhou2021r}&2021&M&0.114&0.815&4.712&0.193&0.876&0.959&0.981&3.5M\\
				R-MSFM6~\cite{zhou2021r}&2021&M&0.112&0.806&4.704&0.191&0.878&0.960&0.981&3.8M\\
				MonoFormer~\cite{bae2023deep}&2023&M&0.108&0.806&4.594&0.184&0.884&\underline{0.963}&\underline{0.983}&23.9M+\\
				SRDepth-Res18~\cite{liu2023self}&2023&M&0.111&\underline{0.762}&4.619&0.186&0.877&0.961&\underline{0.983}&23.3M\\		
				Lite-Mono~\cite{zhang2023lite}&2023&M&\underline{0.107} &   0.765  &   \underline{4.561} &   \underline{0.183}  &   \underline{0.886}  &   \underline{0.963}  &  \underline{0.983}  &
				3.1M \\	
				\textbf{MonoDiffusion (ours)}&2023&M&\textbf{0.103}  &   \textbf{0.726}  &   \textbf{4.447}  &   \textbf{0.179}  &   \textbf{0.893}  &   \textbf{0.965}  &   \textbf{0.984}  & 3.1M \\ 			
				\cline{1-11}
				\hline			
				Monodepth2-Res18~\cite{godard2019digging}&2019&M\dag&0.132&1.044&5.142&0.210&0.845&0.948&0.977&14.3M\\
				Monodepth2-Res50~\cite{godard2019digging}&2019&M\dag&0.131&1.023&5.064&0.206&0.849&0.951&0.979&32.5M\\
				R-MSFM3~\cite{zhou2021r}&2021&M\dag&0.128&0.965&5.019&0.207&0.853&0.951&0.977&3.5M\\
				R-MSFM6~\cite{zhou2021r}&2021&M\dag&0.126&0.944&4.981&0.204&0.857&0.952&0.978&3.8M\\
				Lite-Mono~\cite{zhang2023lite}&2023&M\dag&\underline{0.121}  &  \underline{0.876}  &  \underline{4.918}  &  \underline{0.199}  &   \textbf{0.859}  &  \underline{0.953}  &  \underline{0.980}&3.1M \\
				\textbf{MonoDiffusion (ours)}&2023&M\dag&\textbf{0.119}  &   \textbf{0.843}  &   \textbf{4.868}  &   \textbf{0.196}  &  \underline{0.858}  &   \textbf{0.955}  &   \textbf{0.981}  & 3.1M \\ 
				\hline					
				Monodepth2-Res18~\cite{godard2019digging} &2019&M*&0.115&0.882&4.701&0.190&0.879&0.961&0.982&14.3M\\
				R-MSFM3~\cite{zhou2021r}&2021&M*&0.112&0.773&4.581&0.189&0.879&0.960&0.982&3.5M\\
				R-MSFM6~\cite{zhou2021r}&2021&M*&0.108&0.748&4.470&0.185&0.889&0.963&0.982&3.8M\\
				HR-Depth~\cite{lyu2021hr}&2021&M*&0.106&0.755&   4.472&0.181&0.892& 0.966&\textbf{0.984}&14.7M\\
				SRDepth-Res18~\cite{liu2023self}&2023&M*&0.106&\underline{0.673}&4.379&0.180&0.886&0.965&\textbf{0.984}&23.3M\\		
				Lite-Mono~\cite{zhang2023lite} &2023&M*& 0.102  &   0.746  &   4.444  &   0.179  &  0.896  &   0.965  &  \underline{0.983}&3.1M  \\
				Lite-Mono-8M~\cite{zhang2023lite}&2023&M*&\underline{0.097} & 0.710  &   4.309 &   \underline{0.174}  &   \underline{0.905}  &   \underline{0.967}  &   \textbf{0.984}&8.7M\\
				\textbf{MonoDiffusion (ours)}&2023&M*&0.099  &   0.702  &   \underline{4.305}  &   0.175  &   0.903  &   \underline{0.967}  &   \textbf{0.984}  & 3.1M \\ 
				\textbf{MonoDiffusion-8M (ours)}&2023&M*&\textbf{0.094}  &   \textbf{0.662}  &   \textbf{4.235}  &   \textbf{0.171}  &   \textbf{0.908}  &   \textbf{0.968}  &   \textbf{0.984}  & 8.8M \\ 
				\hline
				\hline
				MonoViT-tiny~\cite{zhao2022monovit}&2022&M&0.102&0.733&4.459&\underline{0.177}&0.895&\underline{0.965}&\textbf{0.984}&10.3M\\
				Lite-Mono-8M~\cite{zhang2023lite}& 2023&M&   \underline{0.101}  &   \underline{0.729}  &   \underline{4.454}  &   0.178  &   \underline{0.897}  &   \underline{0.965}  &   \underline{0.983}&8.7M \\
				\textbf{MonoDiffusion-8M (ours)}&2023&M&\textbf{0.099}  &   \textbf{0.692}  &   \textbf{4.377}  &   \textbf{0.175}  &   \textbf{0.899}  &   \textbf{0.966}  &   \textbf{0.984}  & 8.8M \\ 						
				\hline					
		\end{tabular}}
	\end{center}
	\caption{\textbf{Quantitative depth comparison on the KITTI dataset with the Eigen split~\cite{eigen2014depth}.} The default resizing for all input images is set to $640\times 192$, unless otherwise specified. The best and second best results are indicated in \textbf{bold} and \underline{underlined}, respectively. The following abbreviations are used: ``M" stands for KITTI monocular videos, ``M+Se" indicates monocular videos with added semantic segmentation, ``M*" represents input resolution of $1024\times 320$, and ``M$\dag$" denotes backbones without pre-training on ImageNet~\cite{deng2009imagenet}. MonoDiffusion and MonoDiffusion-8M uses Lite-Mono and Lite-Mono-8M as encoders, respectively. Aside from the depth error and accuracy, we report the model size.}
	\label{table2}
\end{table*}

\textbf{DDIM loss}. Following~\cite{song2020denoising}, we apply a DDIM loss to supervise the predicted noise at each denoising step by revsersing the diffusion process,
\begin{equation}{\mathcal{L}_{{\mathop{ddim}\nolimits} }} = \sum\limits_\textbf{p} {{{\left\| {\epsilon - {\epsilon_\theta }\left( {{\textbf{D}_\tau},\tau, {\textbf{c}} } \right)} \right\|}^2}}.
\end{equation}

\textbf{Edge-aware smoothness loss}. As defined in~\cite{godard2019digging}, we apply an edge-aware smoothness loss to encourage the smoothness property of depth map, 
\begin{equation}
	{{\mathcal L}_{es}} =\sum_{\textbf{p}}{\left| { {\nabla {\textbf{D}^t }\left( \textbf{p} \right)}} \right|}  \odot  {\rm{exp}^{ - {\left\| \nabla {{\textbf{f}^t}\left( \textbf{p}\right)} \right\|_1}}},
\end{equation}
where ${\nabla {\textbf{D}^t }\left( \textbf{p} \right)}$ and $\nabla {{\textbf{f}^t}\left( \textbf{p} \right)}$ calculates the first-order gradients of depth map and RGB image, respectively. ~\\

\textbf{Overall loss}. The overall optimization objective is summarized as
\begin{equation}
	{{\mathcal L}_{overall}} = {\lambda _1}{{\mathcal L}_{ph}} + {\lambda _2}{{\mathcal L}_{KD}} + {\lambda _3}{{\mathcal L}_{rec}} + {\lambda _4}{{\mathcal L}_{{\mathop{ddim}\nolimits} }},
\end{equation}
where ${\lambda _1}$, ${\lambda _2}$, ${\lambda _3}$ and ${\lambda _4}$ are empirically set to 1, 1, 0.1 and 1, respectively. Similar to~\cite{godard2019digging}, we employ two source frames in ${\mathcal L}_{ph}$ and the one with the minimum ${\mathcal L}_{ph}$ is chosen to handle occlusion. Moreover, an auto-masking mechanism~\cite{godard2019digging} is used to deal with dynamic objects. 

	\section{Experiment}
	We conduct extensive experiments on two standard datasets, which includes KITTI~\cite{geiger2012we} and Make3D~\cite{saxena2008make3d}. In the following part, we first describe the relevant datasets, evaluation metrics and implementation details. Then, we present quantitative and qualitative comparisons to the state-of-the-art competitors. Finally, we conduct zero-shot generalization and ablation studies to perform a thorough analysis of MonoDiffusion.
	
	\subsection{Datasets and Evaluation Metrics}
	The \textbf{KITTI} dataset is composed of 61 stereo road scenes with the image resolution around $1241 \times 376$ pixels. The data collection involves multiple sensors such as cameras, 3D Lidar, GPU/IMU, among others. We adopt the Eigen split~\cite{eigen2014depth}, which comprises 39180 monocular triplets for training, 4424 images for validation, and 697 images for testing. As in~\cite{godard2019digging}, we adopt the same intrinsic for all images by averaging the intrinsic of each image. During the evaluation phase, the predicted depth is constrained to the common practice range of $\left[ {0,80} \right]$m.
	
	The \textbf{Make3D} dataset consists of 134 test images  collected from outdoor scenes. We only use this dataset for a zero-shot generalization study by utilizing models trained on the KITTI dataset.
	
	\subsection{Evaluation Metrics}
	Similar to~\cite{zhang2023lite}, we employ the following  evaluation metrics in our experiments,
	\begin{itemize}
		\item Abs Rel: $\frac{1}{\left\| {{\bm{{\rm D}}}} \right\|_1}\sum\nolimits_{{{{ d}}} \in {\bm{{\rm D}}}} {\left| {{{{ d_{t}}}} - {{{d}}}} \right|} /{{{ d}}}$; 
		\item Sq Rel: $\frac{1}{\left\| {{\bm{{\rm D}}}} \right\|_1}{\sum\nolimits_{ {{{ d}}} \in {\bm{{\rm D}}}} {\left\| { {{{ d_t}}} - {{{ d}}}} \right\|} ^2}/{{{ d}}}$;
		\item RMSE: $\sqrt {\frac{1}{\left\| {{\bm{{\rm D}}}} \right\|_1}{{\sum\nolimits_{ {{{ d}}} \in {\bm{{\rm D}}}} {\left\| { {{{ d_t}}} - {{{ d}}}} \right\|} }^2}}$;
		\item RMSE log: $\sqrt {\frac{1}{\left\| {{\bm{{\rm D}}}} \right\|_1}{{\sum\nolimits_{ {{{ d}}} \in {\bm{{\rm D}}}} {\left\| { \log {{{ d_t}}} - \log{{{d}}}} \right\|} }^2}}$;
		\item $\delta < thr$: $\%$ of ${{{{ d}}}}$ satisfies $\left( {\max \left( {\frac{{{{ {{{ d_t}}}}}}}{{{{{{ d}}}}}},\frac{{{{{{ d}}}}}}{{{{ {{{ d_t}}}}}}}} \right) = \delta  < thr} \right)$ for $thr = 1.25,{1.25^2},{1.25^3}.$ 
	\end{itemize}
    Abs Rel, Sq Rel, RMSE and RMSE log are depth error metrics and the lower the better. $\delta < 1.25$, $\delta < 1.25^2$ and $\delta < 1.25^3$ are depth accuracy metrics and the higher the better. Because the self-supervised MDE system has inherent scale ambiguity, we make use of the median scaling technique introduced by~\cite{zhou2017unsupervised} to recover the absolute scale for depth evaluation.

\begin{figure*}[!]
		\centering
		\includegraphics[width=1.0\linewidth]{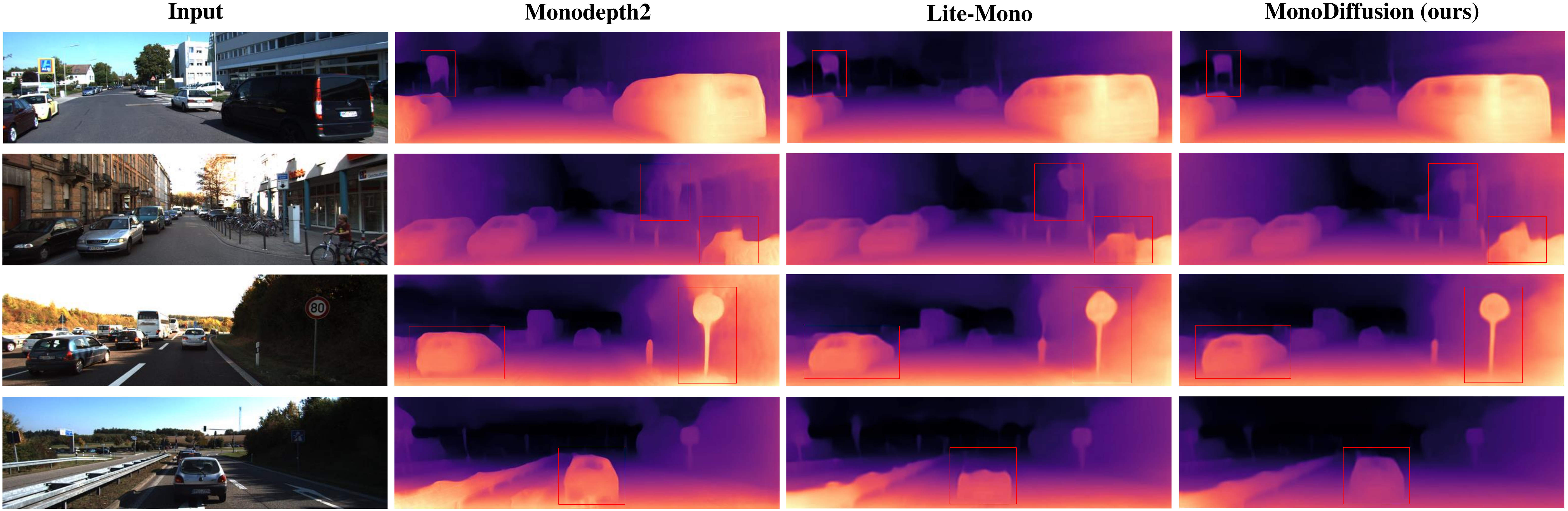}
		\caption{\textbf{Qualitative depth comparison on the KITTI dataset.} The red boxes indicate the regions to emphasize.}
		\label{Fig3}
	\end{figure*}
\begin{figure*}[!]
	\centering
	\includegraphics[width=1.0\linewidth]{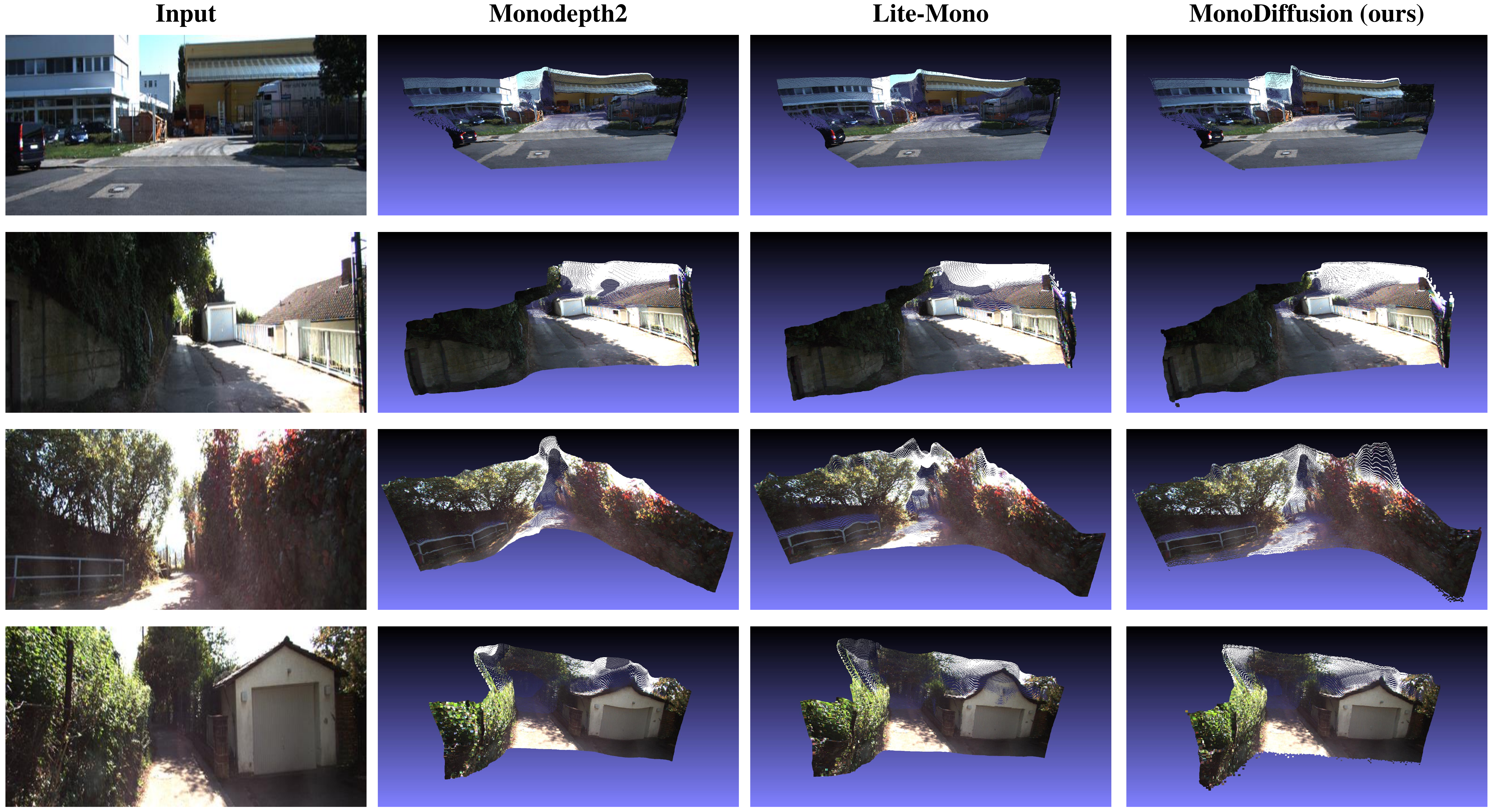}
	\caption{\textbf{Qualitative point cloud results on the KITTI dataset.}}
	\label{Fig7}
\end{figure*}
	
	\subsection{Implementation Details}
	MonoDiffusion is implemented in the PyTorch library and trained on a single NVIDIA TITAN RTX. We use the AdamW optimizer~\cite{loshchilov2017decoupled} where the weight decay is 1e-2 and the batch size is set to 12. When training models from scratch, an initial learning rate of 5e-4 is adopted along with a cosine learning rate schedule~\cite{loshchilov2016sgdr}. The training process runs a total number of 35 epochs in this scenario. It is observed that pre-training on the ImageNet~\cite{deng2009imagenet} accelerates network convergence, resulting in a shorter training time of 30 epochs when loading the pre-trained weights. Moreover, the initial learning rate is adjusted to 1e-4. Following~\cite{zhang2023lite,bae2023deep}, we leverage random horizontal flip as well as random brightness adjustment, saturation adjustment, contrast adjustment and hue jitter with a 50$\%$ chance. We utilize the improved sampling process with 1000 diffusion steps for training and 20 denoising steps for inference.
	
		\begin{table}[tbh]
		\begin{center}
				\scalebox{0.95}{
					\begin{tabular}{c|cccc}
						\hline
						Method & Abs Rel $\downarrow$ & Sq Rel $\downarrow$& RMSE $\downarrow$& RMSE log$\downarrow$ \\
						\hline
						\hline	
						Monodepth~\cite{godard2017unsupervised}&0.544&10.940&11.760&0.193\\
						SfMLearner~\cite{zhou2017unsupervised}&0.383&5.321&10.470&0.478\\
						DDVO~\cite{wang2018learning}&0.387&4.720&8.090&0.204\\
						Monodepth2~\cite{godard2019digging}&0.322&3.589&7.417&0.163\\
						R-MSFM6~\cite{zhou2021r}&0.334&3.285&7.212&0.169\\
						Lite-Mono~\cite{zhang2023lite} &0.305  &  3.060  &  6.981  &  0.158\\
						\textbf{MonoDiffusion (ours)} &\textbf{0.295}&\textbf{2.849}&\textbf{6.854}&\textbf{0.150}\\
						\hline
				\end{tabular}}
			\end{center}
			\caption{\textbf{Quantitative comparison on the Make3D~\cite{saxena2008make3d} dataset.} All models are trained on KITTI~\cite{geiger2012we} with an image resolution of $640\times 192$.}
			\label{table3}
		\end{table}
	\begin{figure*}[!]
		\centering
		\includegraphics[width=1.0\linewidth]{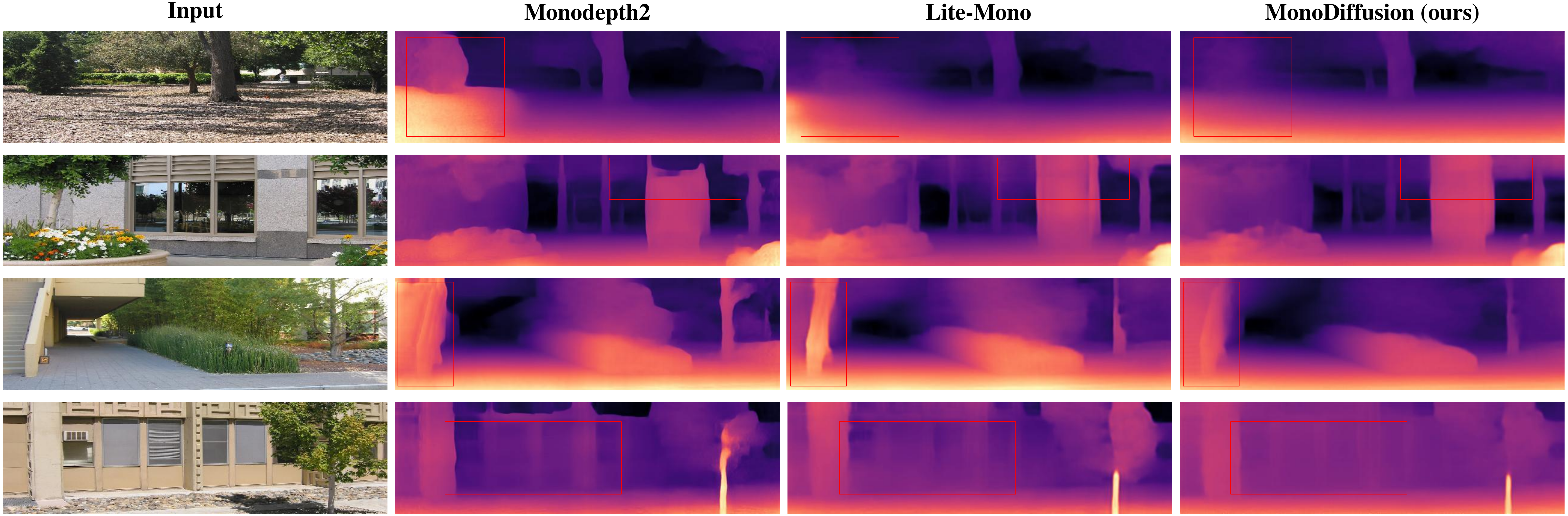}
		\caption{\textbf{Qualitative depth comparison on the Make3D dataset}. }
		\label{Fig4}
	\end{figure*}
	\begin{figure*}[!]
		\centering
		\includegraphics[width=1.0\linewidth]{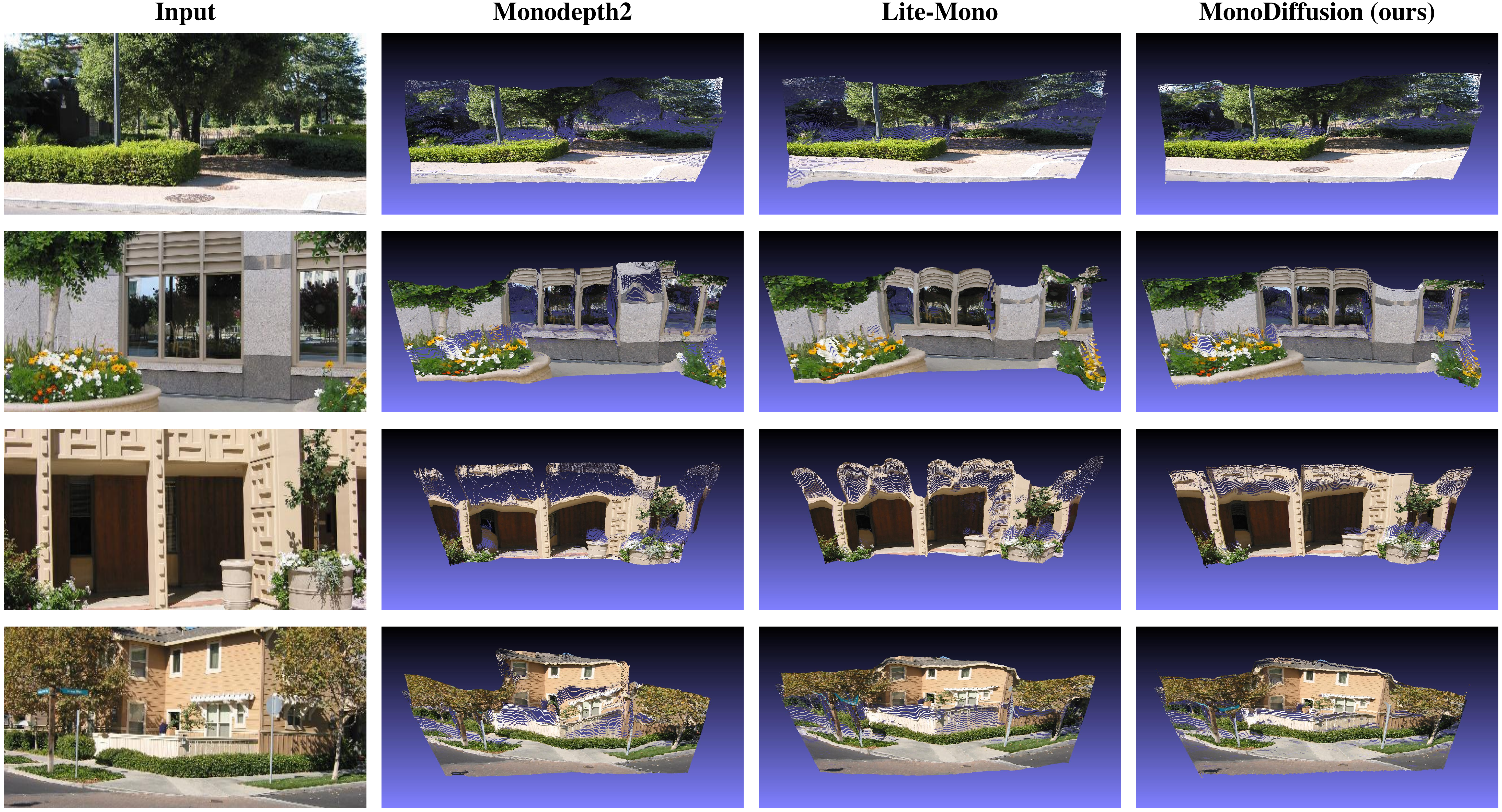}
		\caption{\textbf{Qualitative point cloud results on the Make3D dataset.}}
		\label{Fig8}
	\end{figure*}
	\subsection{Comparison to Prior State-of-the-Arts}
	
		\begin{table*}[htb!]
		\begin{center}
			\renewcommand{\arraystretch}{1.3}
			\resizebox{1.75\columnwidth}{!}{\begin{tabular}{c c  c c c c c c c c c c }
					\Xhline{1.2pt}
					ID & SD & PGD & $\mathcal{L}_{KD}$ & MVC& Abs Rel $\downarrow$ & Sq Rel $\downarrow$ & RMSE $\downarrow$& RMSE log $\downarrow$ & $\delta  < 1.25$ $\uparrow$ &$\delta <1.25^2$ $\uparrow$ &$\delta <1.25^3$ $\uparrow$ \\
					\hline
					\hline				
					1&&& && 0.107 & 0.787 & 4.580 & 0.182&0.887&0.963&0.983\\
					2 &\cmark&&&& 0.434 & 4.630 & 11.99 & 0.579&0.306&0.568&0.777\\	
					3 &&\cmark&&& 0.105 & 0.757 & 4.521 & 0.181&0.889&0.964&\textbf{0.984}\\	
					4 &&\cmark&\cmark&& 0.104 & 0.741 & 4.489 & 0.180&0.890&0.964&\textbf{0.984}\\	
					5 &&\cmark&\cmark&\cmark& \textbf{0.103}& \textbf{0.726} & \textbf{4.447} & \textbf{0.179}&\textbf{0.893}&\textbf{0.965}&\textbf{0.984} \\
					\Xhline{1.2pt}			
			\end{tabular}}
		\end{center}
		\caption{\textbf{Ablation study on the MonoDiffusion}. SD: self-diffusion~\cite{duan2023diffusiondepth}; PGD: pseudo ground-truth diffusion; MVC: masked visual condition mechanism.}
		\label{table4}
	\end{table*}
	
	\begin{table*}[htb!]
		\begin{center}
			\renewcommand{\arraystretch}{1.3}
			\resizebox{1.5\columnwidth}{!}{\begin{tabular}{c|| c c c c c c c  }	
					\Xhline{1.2pt}
					Inference step& Abs Rel $\downarrow$ & Sq Rel $\downarrow$ & RMSE $\downarrow$& RMSE log $\downarrow$ & $\delta  < 1.25$ $\uparrow$ & $\delta  < 1.25^2$ $\uparrow$ & $\delta  < 1.25^3$ $\uparrow$\\
					\hline						
					\hline
					2 &1.812&75.600&30.791&6.160&0.070&0.141&0.213
					\\
					5&1.503&58.013&27.140&5.464&0.125&0.241&0.340
					\\
					10&0.226&3.730&9.571&0.754&0.703&0.878&0.938
					\\
					15 &0.822 &28.099&19.232&2.972&0.357&0.548&0.653
					\\
					20 & \textbf{0.103}& \textbf{0.726} & \textbf{4.447} & \textbf{0.179}&\textbf{0.893}&\textbf{0.965}&\textbf{0.984}
					\\
					25 &0.431 &10.382&12.709&1.206&0.503&0.723&0.827
					\\
					\Xhline{1.2pt}		
			\end{tabular}}
		\end{center}
		\caption{\textbf{Ablation study on different inference steps.}
		}
		\label{table5}
	\end{table*}
	
	\begin{figure}[!]
		\centering
		\includegraphics[width=0.95\linewidth]{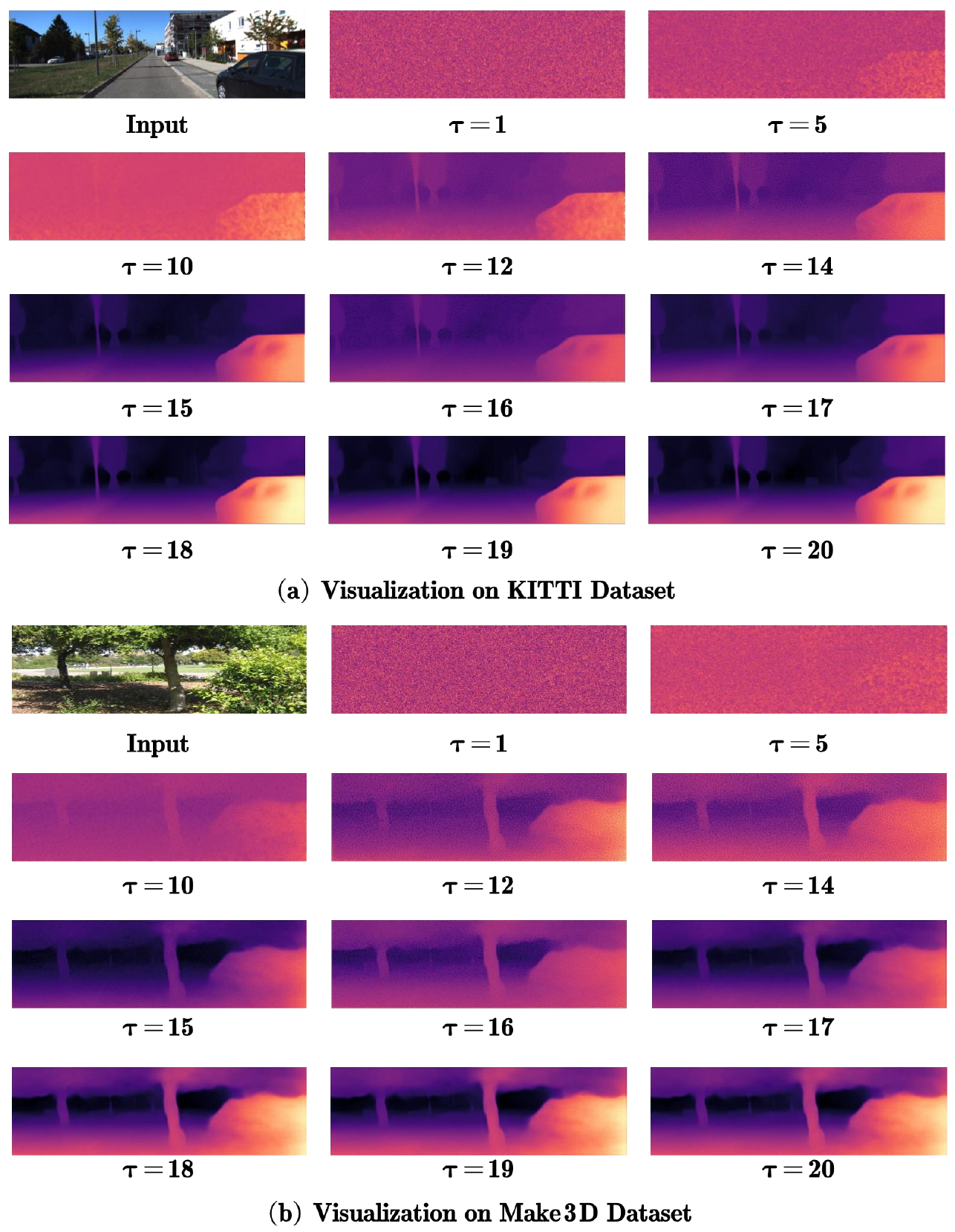}
		\caption{\textbf{Visualization of the denoising process involving 20 inference steps in total.} }
		\label{Fig5}
	\end{figure}

	We compare the proposed MonoDiffusion with prior state-of-the-art competitors on the Eigen split of KITTI benchmark. We mainly focus on training with monocular videos in three settings: $640 \times 192$ resolution, without pre-training on ImageNet and $1024 \times 320$ resolution. The results are summarized in Table~\ref{table2}. As we can see, MonoDiffusion is able to exceed the compared methods in each setting and beats recent Lite-Mono with almost identical number of parameters. Moreover, even compared to semantic segmentation-assisted methods such as SGDepth, MonoDiffusion shows great advantages. 
	
	In Fig.~\ref{Fig3}, we show a qualitative depth comparison. MonoDiffusion is better at delineating object contours, for example, traffic sign and preserves fine-grained depth details. To further show the strengths of MonoDiffusion, we convert depth maps into point clouds and display the 3D structures in Fig~\ref{Fig7}. It can be seen that MonoDiffusion is capable of recovering the 3D world reasonably and shows less distortion than the compared methods. The outstanding performance of our MonoDiffusion evidences the strong generative capability of diffusion model.
		 
	\subsection{Zero-shot Generalization}
	In Table~\ref{table3}, we verify the generalization ability of MonoDiffusion in a fully zero-shot setting. The models are trained on the KITTI dataset but evaluated on the Make3D dataset. The superior results of MonoDiffusion indicate that it generalizes well to unseen scenarios. In Fig.~\ref{Fig4}, we demonstrate a qualitative depth comparison. As we can see, MonoDiffusion acquires highly-detailed depth maps with good visual quality. In Fig~\ref{Fig8}, we further show a qualitative point cloud comparison. The compared methods struggle with thin structures,~\textit{e.g.}, pole, while MonoDiffusion is able to recover these smaller details and preserves prominent geometric features of the 3D scenes at the same time.
	
	\subsection{Ablation Study}
	To better understand the influence of different components in MonoDiffusion on performance, we provide detailed ablation results.
	
	\textbf{MonoDiffusion} (Table~\ref{table4}). We leverage the Lite-Mono~\cite{zhang2023lite} as our baseline (ID 1). The implementation of diffusion is hard for self-supervised MDE due to the lack of depth ground-truth. To allow the diffusion, we utilize the self-diffusion~\cite{duan2023diffusiondepth} and the proposed pseudo ground-truth diffusion, respectively and find that the self-diffusion does not even make the model converge in the self-supervised setting (ID 2). By contrast, the pseudo ground-truth diffusion enables the model to achieve good performance (ID 3). We further appending the knowledge distillation loss in the training phase and achieves consistent improvements on almost all metrics (ID 4). Finally, we integrate the masked visual condition mechanism and acquire the best results (ID 5).
	
	\textbf{Denoising inference.}  We further explore the properties of different inference steps during the denoising process. It can be seen from Table~\ref{table5} that when the number of steps increases, the performance gradually improves. However, the improvement is not always continuous. For example, the performance of step 10 is better than that of step 15. Besides, when continuing to increase the number of steps from the standard denoising step 20, the performance degrades. 
	
	In Fig.~\ref{Fig5}, we show an intuitive illustration of how the depth is iteratively refined from a random depth distribution. As can be seen, the process begins by initializing shapes and edges of objects and then gradually refines the depth values and rectifies the inter- and intra-object distance relationships. 
	
		\begin{table*}[htb!]
		\begin{center}
			\renewcommand{\arraystretch}{1.3}
			\resizebox{1.5\columnwidth}{!}{\begin{tabular}{c|| c c c c c c c   }	
					\Xhline{1.2pt}
					Mask ratio& Abs Rel $\downarrow$ & Sq Rel $\downarrow$ & RMSE $\downarrow$& RMSE log $\downarrow$ & $\delta  < 1.25$ $\uparrow$ & $\delta  < 1.25^2$ $\uparrow$ & $\delta  < 1.25^3$ $\uparrow$\\
					\hline						
					\hline
					0$\%$ &0.104&0.741&4.489&0.180&0.890&0.964&\textbf{0.984}
					\\ 
					10$\%$&0.104&0.729&4.482&\textbf{0.179}&0.891&0.964&0.983
					\\
					20$\%$& \textbf{0.103}& 0.726 & \textbf{4.447} & \textbf{0.179}&\textbf{0.893}&\textbf{0.965}&\textbf{0.984}
					\\
					40$\%$&\textbf{0.103}&0.741&4.463&\textbf{0.179}&0.892&\textbf{0.965}&\textbf{0.984}				
					\\
					60$\%$ &\textbf{0.103} &\textbf{0.718}&4.462&\textbf{0.179}&0.891&\textbf{0.965}&\textbf{0.984}
					\\
					80$\%$ &0.107 &0.772&4.587&0.182&0.885&0.963&0.983
					\\
					\Xhline{1.2pt}		
			\end{tabular}}
		\end{center}
		\caption{\textbf{Ablation study on different mask ratios in the masked visual condition mechanism,} which is only utilized during the training phase.
		}
		\label{table6}
	\end{table*}
	
	\begin{figure}[!]
		\centering
		\includegraphics[width=1.0\linewidth]{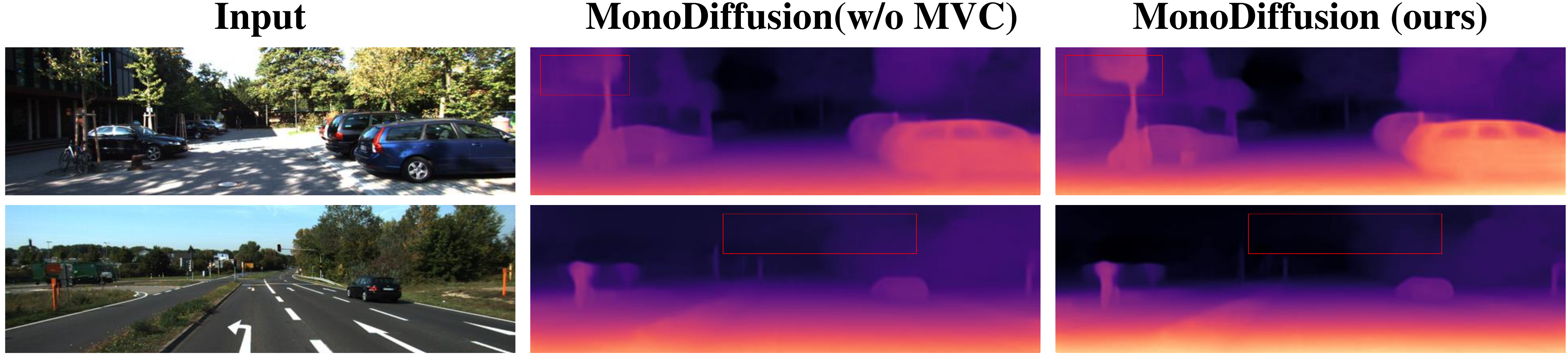}
		\caption{\textbf{Qualitative comparison between MonoDiffusion with and without masked visual condition (MVC) mechanism.} }
		\label{Fig6}
	\end{figure}

	\textbf{Masked visual condition.} In Fig.~\ref{Fig6}, we display a qualitative depth comparison between MonoDiffusion with and without masked visual condition (MVC) mechanism. As can be seen, the MVC mechanism can indeed enhance the denoising ability of model,~\textit{e.g.}, better tree silhouettes. Additionally, the depth at distant distances can be better recovered, for instance, the sky. In the absence of MVC mechanism, the depth of sky is similar to the depth of distant trees, but in fact their depth should be highly different. The reason behind this  may be that using the masked visual conditions to reconstruct the denoised depth enables the noise predictor to better exploit contextual information. As presented in Table~\ref{table6}, we further explore the impact of different mask ratios in the MVC mechanism. We find that 20$\%$ is the best choice, and the model is capable of achieving feasible results even with a mask training ratio of 80$\%$.
	
	\section{Conclusion}
	
	In this work, we develop a novel self-supervised monocular depth estimation framework by reformulating it as an iterative denoising process. Besides, we introduce the pseudo ground-truth diffusion process to assist the diffusion and the masked visual condition mechanism to strengthen the denoising ability of model. Extensive experiments on the KITTI and Make3D datasets indicate the efficacy of the proposed MonoDiffusion. We hope our novel initiative will encourage more sophisticated diffusion-based
	depth estimation achievements.

\normalem
{\small
	\bibliographystyle{unsrt}
	\bibliography{egbib}

\begin{thebibliography}{10}

\bibitem{izadi2011kinectfusion}
Shahram Izadi, David Kim, Otmar Hilliges, David Molyneaux, Richard Newcombe,
  Pushmeet Kohli, Jamie Shotton, Steve Hodges, Dustin Freeman, Andrew Davison,
  et~al.
\newblock Kinectfusion: real-time 3d reconstruction and interaction using a
  moving depth camera.
\newblock In {\em Proceedings of the 24th annual ACM symposium on User
  interface software and technology}, pages 559--568, 2011.

\bibitem{chen2019towards}
Po-Yi Chen, Alexander~H Liu, Yen-Cheng Liu, and Yu-Chiang~Frank Wang.
\newblock Towards scene understanding: Unsupervised monocular depth estimation
  with semantic-aware representation.
\newblock In {\em Proceedings of the IEEE Conference on Computer Vision and
  Pattern Recognition}, pages 2624--2632, 2019.

\bibitem{natan2022end}
Oskar Natan and Jun Miura.
\newblock End-to-end autonomous driving with semantic depth cloud mapping and
  multi-agent.
\newblock {\em IEEE Transactions on Intelligent Vehicles}, 2022.

\bibitem{eigen2014depth}
David Eigen, Christian Puhrsch, and Rob Fergus.
\newblock Depth map prediction from a single image using a multi-scale deep
  network.
\newblock In {\em Advances in Neural Information Processing Systems}, pages
  2366--2374, 2014.

\bibitem{zhou2017unsupervised}
Tinghui Zhou, Matthew Brown, Noah Snavely, and David~G Lowe.
\newblock Unsupervised learning of depth and ego-motion from video.
\newblock In {\em Proceedings of the IEEE Conference on Computer Vision and
  Pattern Recognition}, pages 1851--1858, 2017.

\bibitem{yin2019enforcing}
Wei Yin, Yifan Liu, Chunhua Shen, and Youliang Yan.
\newblock Enforcing geometric constraints of virtual normal for depth
  prediction.
\newblock In {\em Proceedings of the IEEE International Conference on Computer
  Vision}, pages 5684--5693, 2019.

\bibitem{wang2023planedepth}
Ruoyu Wang, Zehao Yu, and Shenghua Gao.
\newblock Planedepth: Self-supervised depth estimation via orthogonal planes.
\newblock In {\em Proceedings of the IEEE Conference on Computer Vision and
  Pattern Recognition}, pages 21425--21434, 2023.

\bibitem{lee2019big}
Jin~Han Lee, Myung-Kyu Han, Dong~Wook Ko, and Il~Hong Suh.
\newblock From big to small: Multi-scale local planar guidance for monocular
  depth estimation.
\newblock {\em arXiv preprint arXiv:1907.10326}, 2019.

\bibitem{bhat2021adabins}
Shariq~Farooq Bhat, Ibraheem Alhashim, and Peter Wonka.
\newblock Adabins: Depth estimation using adaptive bins.
\newblock In {\em Proceedings of the IEEE Conference on Computer Vision and
  Pattern Recognition}, pages 4009--4018, 2021.

\bibitem{kim2018deep}
Youngjung Kim, Hyungjoo Jung, Dongbo Min, and Kwanghoon Sohn.
\newblock Deep monocular depth estimation via integration of global and local
  predictions.
\newblock {\em IEEE Transactions on Image Processing}, 27(8):4131--4144, 2018.

\bibitem{shao2023urcdc}
Shuwei Shao, Zhongcai Pei, Weihai Chen, Ran Li, Zhong Liu, and Zhengguo Li.
\newblock Urcdc-depth: Uncertainty rectified cross-distillation with cutflip
  for monocular depth estimation.
\newblock {\em arXiv preprint arXiv:2302.08149}, 2023.

\bibitem{ye2021unsupervised}
Xinchen Ye, Xin Fan, Mingliang Zhang, Rui Xu, and Wei Zhong.
\newblock Unsupervised monocular depth estimation via recursive stereo
  distillation.
\newblock {\em IEEE Transactions on Image Processing}, 30:4492--4504, 2021.

\bibitem{peng2021excavating}
Rui Peng, Ronggang Wang, Yawen Lai, Luyang Tang, and Yangang Cai.
\newblock Excavating the potential capacity of self-supervised monocular depth
  estimation.
\newblock In {\em Proceedings of the IEEE International Conference on Computer
  Vision}, pages 15560--15569, 2021.

\bibitem{karpov2022exploring}
Aleksei Karpov and Ilya Makarov.
\newblock Exploring efficiency of vision transformers for self-supervised
  monocular depth estimation.
\newblock In {\em 2022 IEEE International Symposium on Mixed and Augmented
  Reality (ISMAR)}, pages 711--719. IEEE, 2022.

\bibitem{godard2019digging}
Cl{\'e}ment Godard, Oisin Mac~Aodha, Michael Firman, and Gabriel~J Brostow.
\newblock Digging into self-supervised monocular depth estimation.
\newblock In {\em Proceedings of the IEEE International Conference on Computer
  Vision}, pages 3828--3838, 2019.

\bibitem{jung2021fine}
Hyunyoung Jung, Eunhyeok Park, and Sungjoo Yoo.
\newblock Fine-grained semantics-aware representation enhancement for
  self-supervised monocular depth estimation.
\newblock In {\em Proceedings of the IEEE International Conference on Computer
  Vision}, pages 12642--12652, 2021.

\bibitem{klingner2020self}
Marvin Klingner, Jan-Aike Term{\"o}hlen, Jonas Mikolajczyk, and Tim
  Fingscheidt.
\newblock Self-supervised monocular depth estimation: Solving the dynamic
  object problem by semantic guidance.
\newblock In {\em European Conference on Computer Vision}, pages 582--600.
  Springer, 2020.

\bibitem{zhang2023lite}
Ning Zhang, Francesco Nex, George Vosselman, and Norman Kerle.
\newblock Lite-mono: A lightweight cnn and transformer architecture for
  self-supervised monocular depth estimation.
\newblock In {\em Proceedings of the IEEE Conference on Computer Vision and
  Pattern Recognition}, pages 18537--18546, 2023.

\bibitem{hoogeboom2022equivariant}
Emiel Hoogeboom, V{\i}ctor~Garcia Satorras, Cl{\'e}ment Vignac, and Max
  Welling.
\newblock Equivariant diffusion for molecule generation in 3d.
\newblock In {\em International Conference on Machine Learning}, pages
  8867--8887. PMLR, 2022.

\bibitem{trippe2022diffusion}
Brian~L Trippe, Jason Yim, Doug Tischer, Tamara Broderick, David Baker, Regina
  Barzilay, and Tommi Jaakkola.
\newblock Diffusion probabilistic modeling of protein backbones in 3d for the
  motif-scaffolding problem.
\newblock {\em arXiv preprint arXiv:2206.04119}, 2022.

\bibitem{chen2022diffusiondet}
Shoufa Chen, Peize Sun, Yibing Song, and Ping Luo.
\newblock Diffusiondet: Diffusion model for object detection.
\newblock {\em arXiv preprint arXiv:2211.09788}, 2022.

\bibitem{chen2022generalist}
Ting Chen, Lala Li, Saurabh Saxena, Geoffrey Hinton, and David~J Fleet.
\newblock A generalist framework for panoptic segmentation of images and
  videos.
\newblock {\em arXiv preprint arXiv:2210.06366}, 2022.

\bibitem{saxena2023monocular}
Saurabh Saxena, Abhishek Kar, Mohammad Norouzi, and David~J Fleet.
\newblock Monocular depth estimation using diffusion models.
\newblock {\em arXiv preprint arXiv:2302.14816}, 2023.

\bibitem{duan2023diffusiondepth}
Yiqun Duan, Xianda Guo, and Zheng Zhu.
\newblock Diffusiondepth: Diffusion denoising approach for monocular depth
  estimation.
\newblock {\em arXiv preprint arXiv:2303.05021}, 2023.

\bibitem{liu2023self}
Zhong Liu, Ran Li, Shuwei Shao, Xingming Wu, and Weihai Chen.
\newblock Self-supervised monocular depth estimation with self-reference
  distillation and disparity offset refinement.
\newblock {\em IEEE Transactions on Circuits and Systems for Video Technology},
  2023.

\bibitem{gao2023masked}
Shanghua Gao, Pan Zhou, Ming-Ming Cheng, and Shuicheng Yan.
\newblock Masked diffusion transformer is a strong image synthesizer.
\newblock {\em arXiv preprint arXiv:2303.14389}, 2023.

\bibitem{wei2023diffusion}
Chen Wei, Karttikeya Mangalam, Po-Yao Huang, Yanghao Li, Haoqi Fan, Hu~Xu,
  Huiyu Wang, Cihang Xie, Alan Yuille, and Christoph Feichtenhofer.
\newblock Diffusion models as masked autoencoders.
\newblock {\em arXiv preprint arXiv:2304.03283}, 2023.

\bibitem{geiger2012we}
Andreas Geiger, Philip Lenz, and Raquel Urtasun.
\newblock Are we ready for autonomous driving? the kitti vision benchmark
  suite.
\newblock In {\em Proceedings of the IEEE Conference on Computer Vision and
  Pattern Recognition}, pages 3354--3361, 2012.

\bibitem{saxena2008make3d}
Ashutosh Saxena, Min Sun, and Andrew~Y Ng.
\newblock Make3d: Learning 3d scene structure from a single still image.
\newblock {\em IEEE Transactions on Pattern Analysis and Machine Intelligence},
  31(5):824--840, 2008.

\bibitem{laina2016deeper}
Iro Laina, Christian Rupprecht, Vasileios Belagiannis, Federico Tombari, and
  Nassir Navab.
\newblock Deeper depth prediction with fully convolutional residual networks.
\newblock In {\em 2016 Fourth International Conference on 3D Vision}, pages
  239--248. IEEE, 2016.

\bibitem{he2016deep}
Kaiming He, Xiangyu Zhang, Shaoqing Ren, and Jian Sun.
\newblock Deep residual learning for image recognition.
\newblock In {\em Proceedings of the IEEE Conference on Computer Vision and
  Pattern Recognition}, pages 770--778, 2016.

\bibitem{cao2017estimating}
Yuanzhouhan Cao, Zifeng Wu, and Chunhua Shen.
\newblock Estimating depth from monocular images as classification using deep
  fully convolutional residual networks.
\newblock {\em IEEE Transactions on Circuits and Systems for Video Technology},
  28(11):3174--3182, 2017.

\bibitem{fu2018deep}
Huan Fu, Mingming Gong, Chaohui Wang, Kayhan Batmanghelich, and Dacheng Tao.
\newblock Deep ordinal regression network for monocular depth estimation.
\newblock In {\em Proceedings of the IEEE Conference on Computer Vision and
  Pattern Recognition}, pages 2002--2011, 2018.

\bibitem{Yuan_2022_CVPR}
Weihao Yuan, Xiaodong Gu, Zuozhuo Dai, Siyu Zhu, and Ping Tan.
\newblock New crfs: Neural window fully-connected crfs for monocular depth
  estimation.
\newblock {\em Proceedings of the IEEE Conference on Computer Vision and
  Pattern Recognition}, 2022.

\bibitem{liu2023va}
Ce~Liu, Suryansh Kumar, Shuhang Gu, Radu Timofte, and Luc Van~Gool.
\newblock Va-depthnet: A variational approach to single image depth prediction.
\newblock {\em International Conference on Learning Representations}, 2023.

\bibitem{bian2019depth}
Jia-Wang Bian, Zhichao Li, Naiyan Wang, Huangying Zhan, Chunhua Shen, Ming-Ming
  Cheng, and Ian Reid.
\newblock Unsupervised scale-consistent depth and ego-motion learning from
  monocular video.
\newblock In {\em Advances in Neural Information Processing Systems}, 2019.

\bibitem{johnston2020self}
Adrian Johnston and Gustavo Carneiro.
\newblock Self-supervised monocular trained depth estimation using
  self-attention and discrete disparity volume.
\newblock In {\em Proceedings of the IEEE Conference on Computer Vision and
  Pattern Recognition}, pages 4756--4765, 2020.

\bibitem{ho2020denoising}
Jonathan Ho, Ajay Jain, and Pieter Abbeel.
\newblock Denoising diffusion probabilistic models.
\newblock {\em Advances in Neural Information Processing Systems},
  33:6840--6851, 2020.

\bibitem{song2020score}
Yang Song, Jascha Sohl-Dickstein, Diederik~P Kingma, Abhishek Kumar, Stefano
  Ermon, and Ben Poole.
\newblock Score-based generative modeling through stochastic differential
  equations.
\newblock {\em International Conference on Learning Representations}, 2021.

\bibitem{dhariwal2021diffusion}
Prafulla Dhariwal and Alexander Nichol.
\newblock Diffusion models beat gans on image synthesis.
\newblock {\em Advances in Neural Information Processing Systems},
  34:8780--8794, 2021.

\bibitem{song2020denoising}
Jiaming Song, Chenlin Meng, and Stefano Ermon.
\newblock Denoising diffusion implicit models.
\newblock {\em International Conference on Learning Representations}, 2021.

\bibitem{wolleb2022diffusion}
Julia Wolleb, Robin Sandk{\"u}hler, Florentin Bieder, Philippe Valmaggia, and
  Philippe~C Cattin.
\newblock Diffusion models for implicit image segmentation ensembles.
\newblock In {\em International Conference on Medical Imaging with Deep
  Learning}, pages 1336--1348. PMLR, 2022.

\bibitem{brempong2022denoising}
Emmanuel~Asiedu Brempong, Simon Kornblith, Ting Chen, Niki Parmar, Matthias
  Minderer, and Mohammad Norouzi.
\newblock Denoising pretraining for semantic segmentation.
\newblock In {\em Proceedings of the IEEE Conference on Computer Vision and
  Pattern Recognition}, pages 4175--4186, 2022.

\bibitem{baranchuk2021label}
Dmitry Baranchuk, Ivan Rubachev, Andrey Voynov, Valentin Khrulkov, and Artem
  Babenko.
\newblock Label-efficient semantic segmentation with diffusion models.
\newblock {\em International Conference on Learning Representations}, 2022.

\bibitem{sohl2015deep}
Jascha Sohl-Dickstein, Eric Weiss, Niru Maheswaranathan, and Surya Ganguli.
\newblock Deep unsupervised learning using nonequilibrium thermodynamics.
\newblock In {\em International Conference on Machine Learning}, pages
  2256--2265. PMLR, 2015.

\bibitem{jaderberg2015spatial}
Max Jaderberg, Karen Simonyan, Andrew Zisserman, et~al.
\newblock Spatial transformer networks.
\newblock {\em Advances in Neural Information Processing Systems}, 28, 2015.

\bibitem{yang2020d3vo}
Nan Yang, Lukas~von Stumberg, Rui Wang, and Daniel Cremers.
\newblock D3vo: Deep depth, deep pose and deep uncertainty for monocular visual
  odometry.
\newblock In {\em Proceedings of the IEEE Conference on Computer Vision and
  Pattern Recognition}, pages 1281--1292, 2020.

\bibitem{wang2004image}
Zhou Wang, Alan~C Bovik, Hamid~R Sheikh, and Eero~P Simoncelli.
\newblock Image quality assessment: from error visibility to structural
  similarity.
\newblock {\em IEEE Transactions on Image Processing}, 13(4):600--612, 2004.

\bibitem{yin2018geonet}
Zhichao Yin and Jianping Shi.
\newblock Geonet: Unsupervised learning of dense depth, optical flow and camera
  pose.
\newblock In {\em Proceedings of the IEEE Conference on Computer Vision and
  Pattern Recognition}, pages 1983--1992, 2018.

\bibitem{wang2018learning}
Chaoyang Wang, Jos{\'e}~Miguel Buenaposada, Rui Zhu, and Simon Lucey.
\newblock Learning depth from monocular videos using direct methods.
\newblock In {\em Proceedings of the IEEE Conference on Computer Vision and
  Pattern Recognition}, pages 2022--2030, 2018.

\bibitem{yan2021channel}
Jiaxing Yan, Hong Zhao, Penghui Bu, and YuSheng Jin.
\newblock Channel-wise attention-based network for self-supervised monocular
  depth estimation.
\newblock In {\em International Conference on 3D Vision}, pages 464--473. IEEE,
  2021.

\bibitem{lyu2021hr}
Xiaoyang Lyu, Liang Liu, Mengmeng Wang, Xin Kong, Lina Liu, Yong Liu, Xinxin
  Chen, and Yi~Yuan.
\newblock Hr-depth: High resolution self-supervised monocular depth estimation.
\newblock In {\em Proceedings of the AAAI Conference on Artificial
  Intelligence}, volume~35, pages 2294--2301, 2021.

\bibitem{zhou2021r}
Zhongkai Zhou, Xinnan Fan, Pengfei Shi, and Yuanxue Xin.
\newblock R-msfm: Recurrent multi-scale feature modulation for monocular depth
  estimating.
\newblock In {\em Proceedings of the IEEE International Conference on Computer
  Vision}, pages 12777--12786, 2021.

\bibitem{bae2023deep}
Jinwoo Bae, Sungho Moon, and Sunghoon Im.
\newblock Deep digging into the generalization of self-supervised monocular
  depth estimation.
\newblock In {\em Proceedings of the AAAI Conference on Artificial
  Intelligence}, volume~37, pages 187--196, 2023.

\bibitem{zhao2022monovit}
Chaoqiang Zhao, Youmin Zhang, Matteo Poggi, Fabio Tosi, Xianda Guo, Zheng Zhu,
  Guan Huang, Yang Tang, and Stefano Mattoccia.
\newblock Monovit: Self-supervised monocular depth estimation with a vision
  transformer.
\newblock In {\em International Conference on 3D Vision}, pages 668--678. IEEE,
  2022.

\bibitem{deng2009imagenet}
Jia Deng, Wei Dong, Richard Socher, Li-Jia Li, Kai Li, and Li~Fei-Fei.
\newblock Imagenet: A large-scale hierarchical image database.
\newblock In {\em 2009 IEEE conference on computer vision and pattern
  recognition}, pages 248--255. Ieee, 2009.

\bibitem{loshchilov2017decoupled}
Ilya Loshchilov and Frank Hutter.
\newblock Decoupled weight decay regularization.
\newblock {\em International Conference on Learning Representations}, 2018.

\bibitem{loshchilov2016sgdr}
Ilya Loshchilov and Frank Hutter.
\newblock Sgdr: Stochastic gradient descent with warm restarts.
\newblock {\em International Conference on Learning Representations}, 2017.

\bibitem{godard2017unsupervised}
Cl{\'e}ment Godard, Oisin Mac~Aodha, and Gabriel~J Brostow.
\newblock Unsupervised monocular depth estimation with left-right consistency.
\newblock In {\em Proceedings of the IEEE Conference on Computer Vision and
  Pattern Recognition}, pages 270--279, 2017.

\end{thebibliography}
}

\vfill

\end{document}